# k-MS: A novel clustering algorithm based on morphological reconstruction

Érick Oliveira Rodrigues[a],[*], Leonardo Torok[a], Panos Liatsis[b], José Viterbo[a], Aura Conci[a]

[a] *Department of Computer Science, Universidade Federal Fluminense, Niteroi, Rio de Janeiro, Brazil*
[b] *Department of Electrical Engineering, The Petroleum Institute, Abu Dhabi, United Arab Emirates*

**ABSTRACT**

This work proposes a clusterization algorithm called k-Morphological Sets (k-MS), based on morphological reconstruction and heuristics. k-MS is faster than the CPU-parallel k-Means in worst case scenarios and produces enhanced visualizations of the dataset as well as very distinct clusterizations. It is also faster than similar clusterization methods that are sensitive to density and shapes such as Mitosis and TRICLUST. In addition, k-MS is deterministic and has an intrinsic sense of maximal clusters that can be created for a given input sample and input parameters, differing from k-Means and other clusterization algorithms. In other words, given a constant *k*, a structuring element and a dataset, k-MS produces *k* or less clusters without using random/pseudo-random functions. Finally, the proposed algorithm also provides a straightforward means for removing noise from images or datasets in general.

## 1. Introduction

The clusterization or clustering problem can be informally defined as the problem of separating data objects into groups. Objects that are in the same cluster are in general more similar to each other rather than when compared to objects in different clusters. This definition assumes that there is some quality measure responsible for accounting similarity or dissimilarity between objects. Spatial proximity with respect to a given distance can be a possible similarity measure. In summary, clusterization is the problem of aggregating data in different clusters such that the per cluster similarity function is optimized [1].

k-Means is a well known and commonly applied algorithm for solving the clusterization problem. At the very first step, k-Means randomly or pseudo-randomly chooses *k* points that are defined as the means of each cluster. Each instance is then associated with the nearest one of the generated *k* points. At each iteration, the position of each point is updated according to the central or mean position of each cluster, and the instances are then reassigned to their nearest cluster. This process is repeatedly performed until the algorithm converges, given a predefined threshold [2].

The quality of the solution provided by k-Means is closely associated to the initialization step, i.e., the initial locations of the *k* points. Thus, in some cases, it is better to perform several initializations of the k-Means with different positions for the *k* points on each start, and select the best solution. Some works aim to do exactly this, i.e., they improve the quality of the final solution while reducing the processing time required for running the algorithm several times with different initialization conditions [3]. Furthermore, Likas et al. [4] also highlight that the k-Means performance relies heavily on the initial conditions.

The objective of this work is to propose a new clustering algorithm that uses principles of mathematical morphology and borrows some fundamental characteristics of k-Means, such as the use of the *k* variable and the use of a similarity metric. The algorithm proposed in this work produces interesting clustering results as well as a straightforward way of removing noise from datasets. The clusterizations are driven by the input *k* variable, which denotes the maximum number of clusters, as well as by the structuring element. The fact of being able to use different structuring element enables a wealth of possible clusterizations for a single value of *k*. Furthermore, the proposed method can be easily parallelized. For this reason, we explore and extensively compare the performance of parallel and sequential approaches, using GPU and CPU.

The work is organized as follows. In Section 2, we perform a literature review with regards to mathematical morphology, focusing on morphological reconstructions, which is the foundation of the proposed technique. Furthermore, we address some recent works that perform clusterizations with the intent of recognizing shapes and cluster density. However, these works do not employ mathematical morphology, which is more extensively addressed and discussed in Section 2.2. Section 3 describes the proposed methodology, where CPU and GPU approaches are explored and their behavior analysed. We also perform a complexity analysis of the proposed algorithm. In Section 4, we perform several experiments that compare the quality and time performance of k-MS. Finally, Section 5 contains the conclusions of this work, summarizes the features of the proposed algorithm and discusses potential applications.

---

[*] Corresponding author.
*E-mail addresses:* erickr@id.uff.br (É.O. Rodrigues), leotorok@gmail.com (L. Torok), pliatsis@pi.ac.ae (P. Liatsis), viterbo@ic.uff.br (J. Viterbo), aconci@ic.uff.br (A. Conci).

## 2. Literature review

Given a dataset $X = \{x_1, \ldots, x_L\}$, containing $L$ instances, where $x_l \in \mathbb{R}^n$, $0 < l < L \in \mathbb{N}'$, the common clustering problem consists of partitioning $X$ into $k$ disjoint subsets or clusters ($C_1, \ldots, C_k$) such that a cluster criterion is optimized. The most widely used clustering criterion is the sum of the squared Euclidean distances between each instance $x_l$ and its centroid $u_m$ (cluster center) in regards to the subset that contains $x_l$ [4]. This clustering error is defined in Eq. (1):

$$E(C_1, \ldots, C_k) = \sum_{l=1}^{L} \sum_{m=1}^{k} I(x_l \in C_m) \| x_l - u_m \| \tag{1}$$

where $I(Y) = 1$ if $Y$ is true or $I(Y) = 0$, otherwise [4], and $\|\|$ represents the distance norm.

Clustering techniques are applied in a broad range of areas such as in (1) knowledge discovery, e.g., to separate instances that have common characteristics in different categories [2,5,6], in (2) image processing, analysis, pattern recognition and image segmentation, to categorize types of tumours, diseases and abnormalities [7,8] or to infer the content of images [9,10], in (3) adaptive controllers for games [11], where the controller changes shape and adapts in real time according to the input of the user, in (4) robotics [12] for navigation, and in many other instances.

Although k-Means is robust and simple, it does not consider the morphological aspects of the data, which may be very important in some cases such as in image processing. In other words, k-Means is not able to exploit shape and density information. On the contrary, the proposed algorithm overcomes these limitations by using morphological reconstructions, which are sensitive to cluster density and to the morphology itself. In this research, we propose and extend a morphological reconstruction algorithm to include the input parameter $k$.

### 2.1. Mathematical morphology

Mathematical Morphology (MM) usually studies the geometrical structures present in images through mathematics [13]. Its main principle is extracting information related to the geometry and topology of subsets using structuring elements [14]. The structuring element is often defined as a set of elements representing pixels or voxels. Dilation and erosion, which use the input image and a structuring element, are the two principal operations in MM. A binary dilation of an image $A$ by a structuring element $B$ is given by Eq. (2):

$$dil^B(A) = \bigcup_{b \in B} A_b \tag{2}$$

where $A_b$ represents the elements of $A$ translated by $b$, and $b$ represents the elements of set $B$. That is, let us suppose $A = \{(1, 0), (0, 0), (2, 2), (3, 2)\}$ and $b = (1, 1)$, then $A$ translated by $b$ is equal to $A_b = \{(2, 1), (1, 1), (3, 3), (4, 3)\}$. In other words, the dilation operation is the union of $A_b$ for every element $b$ in $B$. Each element in both sets $A$ and $B$ represents the position of a pixel (black or white pixels, according to the adopted convention). Fig. 1 illustrates a dilation of image $A$ by a structuring element $B = \{(0, 0), (0, 1), (0, -1), (1, 0), (-1, 0)\}$. As a remark, erosion is the opposite of dilation, it erases pixels of the image instead of growing them. However, we do not cover erosion in this work since it is not directly related to our approach.

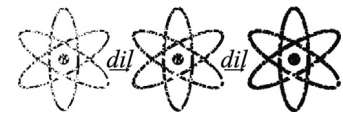

**Fig. 1.** A binary image dilation performed twice.

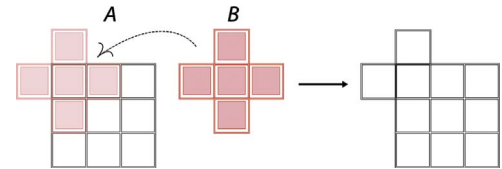

**Fig. 2.** Dilation of a single pixel in $A$.

From an implementation viewpoint, the structuring element is displaced over the input image $A$, where the center of the structuring element corresponds to the iterated pixel in image $A$. At every possible pixel of $A$, the corresponding pixels with regards to the structuring element (highlighted in red) are painted in $A$ if any intersection with the red structuring element occurs, as illustrated in Fig. 2.

It is also possible to perform dilations in grey scale images [15]. In this case, the structuring functions return the value of the pixel for each element of the structuring element. That is, $v()$ returns a natural number for each element $b$ in the structuring element $B$ or $a$ in the image $A$. The grey scale dilation of a single element $a$ in the image $A$ by a structuring element $B$ is then given by Eq. (3):

$$dil^B(a) = sup_{b \in B}[v(a_b) + v(b)] \tag{3}$$

where $sup$ represents the supremum, $a_b$ represents the element $a$ translated by $b$, $v(a_b)$ the grey level of the element $a_b$ in $A$, and $v(b)$ the grey level of the element $b$ in $B$.

From an implementation perspective, the grey scale dilation translates the structuring element over $A$ and computes the sum of the neighbouring pixel values ($a$'s and $b$'s) at each possible position. At every possible position $a$ in $A$, the elements are summed and the supremum of these elements is placed at the iterated pixel of $A$. Fig. 3 shows a grey scale dilation obtained with a circular structuring element (also called ring).

Morphological reconstructions adopt the dilation operation previously described and a mask, which can also be a binary image [16]. The mask operation discards the elements of $A$ that are marked false in the mask. In other words, if repeatedly applied, it limits the dilation to a certain boundary. Given seed points in the image and a mask, we can dilate these points until the image reaches idempotence. Idempotence is defined as applying a function to data twice without changing the result. That is, in our case, idempotence is reached when $A = dil(A) \cap M$, where $M$ is the mask and $A$ is the input image [17].

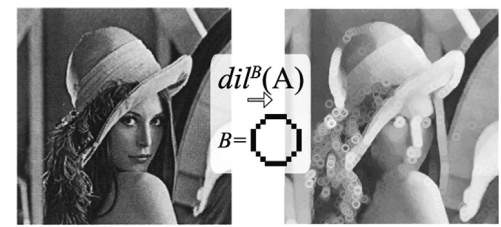

**Fig. 3.** A grey scale dilation using a circular structuring element.

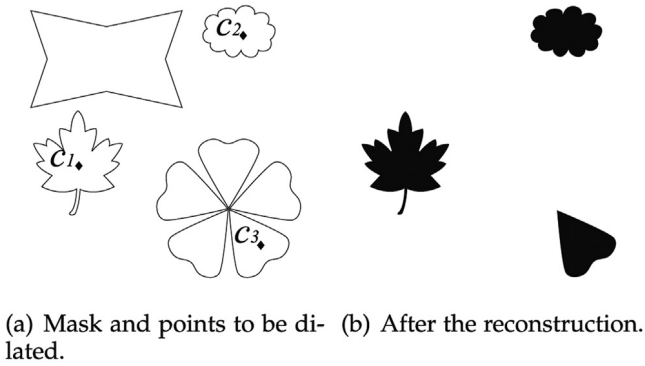

(a) Mask and points to be dilated.   (b) After the reconstruction.

**Fig. 4.** Morphological reconstruction.

Let the lines illustrated in Fig. 4-(a) represent the mask and the points $c_1$, $c_2$ and $c_3$ illustrate the seed points we wish to dilate. After reaching idempotence, the image would look like Fig. 4-(b).

We define the morphological reconstruction using a mask $M$ as $R_M^B(A)$ in Eq. (4),

$$R_M^B(A) = h_Q \qquad (4)$$

where

$$h_{q+1} = dil^B(h_q) \cap M \qquad (5)$$

$$h_0 = A \qquad (6)$$

and $q$ stands for the current iteration until idempotence at iteration $Q$, i.e., until $h_Q \neq h_{Q-1}$.

As it can be seen, the binary morphological reconstruction itself is not a clusterization algorithm. Instead, it just spreads the pixels. The spreading is limited by the mask $M$ and terminates when idempotence is reached. The final result is a binary image, as shown in Fig. 4-(b). However, if we apply the same reconstruction process using a proper structuring element and grey scale images, where each pixel of the grey scale image apart from the background has a unique value, then we start to obtain clusters. In this case, each drawing in Fig. 4-(b) would be represented in a different grey color, that is, they are associated to a unique grey level or index, such as in Fig. 5.

The problem with grey scale morphological reconstructions is that there is no $k$ variable. That is, we cannot limit the amount of clusters we wish to create with a simple morphological reconstruction. Thus, this work specifically proposes an algorithm that fills this gap, i.e., it continuously considers the clusterization result obtained with reconstruction and limits the number of clusters to a maximum of $k$. Moreover, an efficient way of clusters communication, i.e., joining together, is introduced, which is not envisaged in traditional reconstruction. To the best of our knowledge, no work has previously attempted to create a clusterization algorithm using morphological reconstructions and a variable $k$, neither on CPU nor GPU.

### 2.2. Related works

Karypis et al. [18] proposed a clustering algorithm called Chameleon that is capable of indirectly detecting morphology in datasets. However, their algorithm is not based in mathematical morphology itself. Instead, it consists of two phases. In the first phase it uses graph-partitioning to cluster the dataset in relatively small sub-clusters, while in the second phase, it uses an algorithm to find the genuine clusters by repeatedly combining these sub-clusters.

The Chameleon algorithm is categorized as a hierarchical clustering algorithm, which seeks to build an hierarchy of clusters. It can also be categorized as agglomerative instead of divisive because clusters start small and are aggregated to form bigger clusters as time advances, rather than the opposite, i.e., starting with large clusters and dividing them in time.

It is clear that Chameleon algorithm is based on graph operations, which are very different from mathematical morphology. Although their algorithm is able to indirectly detect morphology and shapes, it lacks the richness of configurations that are enabled through the proposed technique. In fact, with k-MS we are able to not only change the $k$ variable, but also the structuring element as well, which alters the fashion in which the clusters propagate. Besides, it is possible to perform pre-processing, such as dilations and erosions or combinations of them, before and after clustering the images. The combination of these aspects makes k-MS very robust and unique.

Liu et al. [19] proposed a clustering algorithm called TRICLUST, which is based on the Delaunay triangulation. Their methodology is substantially different from the mathematical morphology approach proposed in this work. Furthermore, Liu et al. [20] also propose a clustering algorithm that is similar to the previously described approaches, and hence very different in regards to morphological reconstruction. The authors evaluate how well cluster prototypes are separated and form clusters using a separation measure.

Finally, Yousri et al. [21] proposed a clustering algorithm, called Mitosis. This approach also bears similarities to previous works. In this case, Mitosis uses distance relatedness patterns as a measure to identify clusters of different densities. The authors point out some issues of the Chameleon algorithm such as the slow speed of the algorithm and the difficulty in tuning its parameters. These four works are probably the most similar ones in the literature.

In this contribution, these state-of-the-art algorithms are compared and analysed in Section 4. However, we can argue in advance that none of them and, to the extent of our knowledge, no clusterization algorithm [22–26] has the said intrinsic property of the maximum number of clusters that can be found given a certain $k$. Moreover, no other algorithm considers the use of mathematical morphology theory to exploit shape and density information. k-MS is much simpler than the previously described techniques, and in fact, it can be described in a few lines of pseudo-code.

In addition, previous works do not provide sufficient comparisons in terms of run times in the context of a large number of instances. In Yousri et al. [21] and D. Liu et al. [19] a time analysis is given, where they consider a moderate number of instances. However, these analyses do not provide sufficient insight in terms of algorithmic performance. Most of the works provide a complexity analysis of their algorithms, but complexity is not always a fair indicator of speed, as exemplified in Sections 3.1 and 4.3.

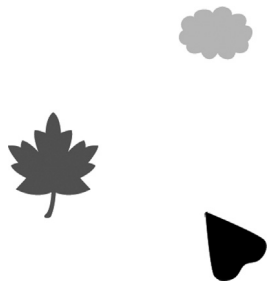

**Fig. 5.** A grey scale morphological reconstruction.

## 3. Proposed methodology

At first, the input dataset to be clusterized must be discretized in order to be processed by the proposed algorithm. The discretization can be done in several distinct ways. One of the ways to perform the required conversion would be to linearly transform the data. Given a dataset $D$ with $L$ instances, such that $D = \{(j_1, i_1), ..., (j_L, i_L)\}$, $D \subset \mathbb{R}^2$, we wish to perform a conversion of this data to a boolean matrix. The $T$ matrix in Eq. (9) stands for a possible boolean matrix such that $f_x(j), f_y(i) \in \mathbb{Z}^2$, where the element at position $(f_x(j), f_y(i))$ in the matrix is equal to 1 ($T_{f_y(i),f_x(j)} = 1$) if $(j, i) \in D$ or is equal to 0 otherwise.

The $f_y$ and $f_x$ represent functions to convert the real attribute values $i$ and $j$ of the instances to natural numbers $f_x(j), f_y(i)$, where both represent the $y$ and $x$ axes of a 2D matrix. The discretization can be done using a simple linear function aggregated to a rounding function such as the ones shown in Eqs. (7) and (8), where $min_i$ and $min_j$ represent the minimum $i$ and $j$, respectively, and $\gamma$ represents a precision factor. For instance, if $i$ equals 32.4, then it may be better to multiply that value by 10 to avoid loss of data, in which case $\gamma$ would be equal to 10. The value 0 can be seen as the image background, and the result is a boolean matrix.

$$f_y(i) = round(\gamma \times (i - min_i)) \quad (7)$$

$$f_x(j) = round(\gamma \times (j - min_j)) \quad (8)$$

$$T_{f_y(i),f_x(j)} = \begin{cases} 1 & if (j, i) \in D \\ 0 & otherwise \end{cases} \quad (9)$$

From the boolean matrix, we then compute another matrix $G$ given by Eq. (10), which is an essential part of the approach. The $w$ variable stands for the width of $G$ and $T$, which are of the same size. In other words, $G$ is a grey scale matrix, where each element has a unique index, expressed in terms of the $x$ and $y$ coordinates, just as in the morphological reconstruction addressed in the previous section. The constant $S$ stands for a value that represents the background (it can set to 0 or any other value as long as $\bar{i}w + \bar{j} \neq S$ for all possible $\bar{i}$ and $\bar{j}$). For convenience, we define that $\bar{i} = f_y(i)$ as well as $\bar{j} = f_x(j)$.

$$G_{\bar{i},\bar{j}} = \begin{cases} \bar{i}w + \bar{j}, & if\ T_{\bar{i},\bar{j}} = 1 \\ S, & otherwise \end{cases} \quad (10)$$

Thus, we now have two matrices $T$ and $G$. $T$ is a boolean matrix, containing 0 s or 1 s, where 0 s represent the background and 1 s represent the instances. $G$, on the other hand, is a matrix that contains a unique natural number on every position of matrix $T$ that contains the value 1. We represent the discretization as well as the algorithm in 2 dimensions and experiment the clusterization proposal using images. However, the idea can be generalized to any number of dimensions in a straightforward fashion.

Once a matrix like $G$ has been built, we start to cluster the data based on its values. An operation similar to a reconstruction $R_M^B(G)$ is applied, where the mask $M$ is equal to $T$. For the experiments in this work, we have used a grey scale structuring element that is equal to $B$, shown in Fig. 6, where all the values are 0 (including the central element marked with ×) and the total size is 1 with respect to the center, regarding the $sup$ metric. However, the size of the structuring element can be changed as desired, which leads to distinct types of clusterizations. In fact, the shape of the structuring element can be changed as long as it contains all directions (e.g., left, right, up and down for the case of 2 dimensions). If this is not the case, the algorithm may not reach idempotence and subsequently it may not converge.

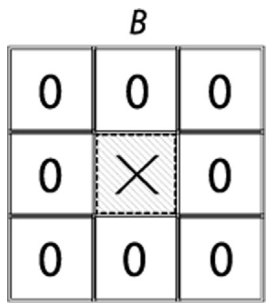

**Fig. 6.** Shape and values of the structuring element $B$.

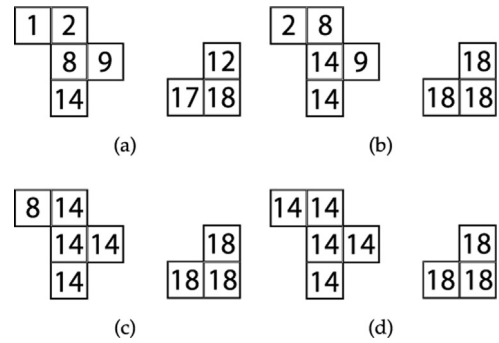

**Fig. 7.** An indexed grey scale morphological reconstruction over 4 iterations.

A structuring element with value 0 implies that the dilation spreads the biggest values through some parts of matrix $G$ (limited by mask $M$), biased by the size and shape of the structuring element. Therefore, after the reconstruction, we obtain a clusterized matrix such as in the process illustrated in Fig. 7, where the image or matrix is being dilated by a structuring element similar to Fig. 2, with zeros as values. In Fig. 7-(a), we have a possible matrix $G$ with grey values {1, 2, 8, 9, 12, 14, 17, 18} (which are also the indexes of the pixels), and all the other pixels are set to $S$ (which are not being shown in the figure). Fig. 7-(a)–(d) illustrate the morphological reconstruction as time advances. In Fig. 7-(d), we have two clusters after a total of 4

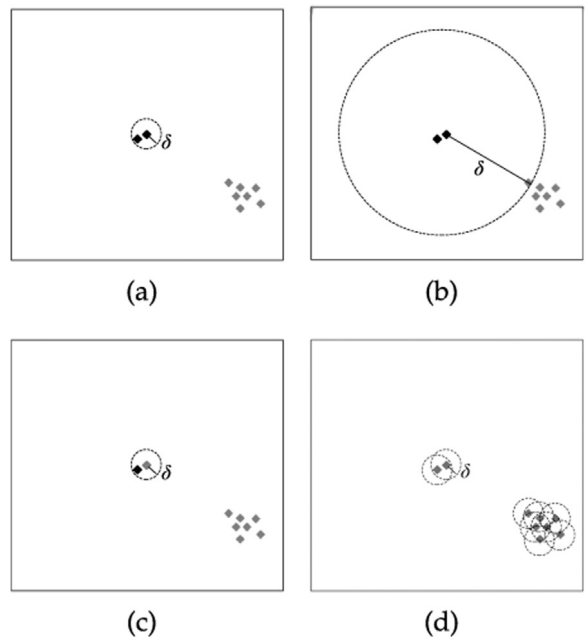

**Fig. 8.** Possible steps of the k-MS algorithm.

iterations (*a* to *d*) and the idempotence has been reached.

As previously addressed, this morphological reconstruction does not separate the data in a prescribed number of clusters. Instead, the algorithm separates the data by "connected" instances based on the size and shape of the structuring element. Therefore, we introduce an internal variable $\delta$, which is responsible for increasing the size of the structuring element $B$. That is, in the beginning of the clusterization, the size of $\delta$ would be minimum, as depicted in Fig. 8-(a). With a structuring element of this size, the algorithm reaches idempotence. Therefore, we gradually increase the size of the structuring element until it reaches the remaining grey instances at the bottom of the matrix in image (b). At each increment of $\delta$, the algorithm checks for the condition of idempotence, and if it is still idempotent, it increments $\delta$. Otherwise, the algorithm resets $\delta$ to the minimum, such as shown in (c). At this moment, the grey level value of the instances at the bottom reaches the upper instances making them belong to the same cluster, as shown in (d). Fig. 8 depicts a hypothetical situation where $k=1$, i.e., just one cluster is desired.

Algorithm 1 shows a high-level overview of the proposed algorithm, where $\delta$ is a factor that multiplies the size of the structuring element $B$. In the beginning of the algorithm, the $\delta$ variable is set to 1. A morphological reconstruction operation is then performed on the image $G$ (see Eq. (10)), using the structuring element $B$, while respecting the mask $T$. The reconstruction consists of dilating the image and applying the mask until idempotence has been reached, as exemplified in Section 2.1 and also in Fig. 7.

**Algorithm 1.** High-level k-Morphological Sets algorithm.

**Data:** $G$ is the grey scale matrix built in Equation 10, $T$ is the mask obtained from Equation 9, $B$ the structuring element, $k$ is the desired maximal number of clusters to be created and $R$ the reconstruction function

1 **begin**
2     $\delta \leftarrow 1$;
3     Perform reconstruction $R_T^{\delta B}(G)$;
4     If the number of clusters in $G$ is less or equal to $k$, i.e., if there is a total number of $k$ or less different grey level values in $G$, stop the algorithm here;
5     $\delta \leftarrow \delta + 1$;
6     Dilate $G$, that is: $G \leftarrow dil^{\delta B}(G)$;
7     If $G$ is idempotent ($G = dil^{\delta B}(G)$) then return to line 5;
8     Otherwise, return to line 2;
9 **end**

Later, the number of unique values in $G$ has to be counted, which is equivalent to the number of clusters present in $G$ at the moment. We count these values while considering early breaks as shown in Algorithm 2 to speed up the processing (lines 10 and 12). If there is more than $k$ clusters in $G$, then the algorithm continues to the next line, where the variable $\delta$ is incremented. $G$ is then dilated once. If $G$ is idempotent, then $\delta$ is incremented until $G$ is not idempotent anymore. The idempotence is overcome when an index that belongs to a certain cluster reaches another with a different index. When this is the case, the algorithm resets $\delta$ to 1 and starts the same process all over again. When there are less than $k$ clusters in $G$, the algorithm converges, as shown in line 4 of Algorithm 1. In contrast to the more general view shown in Algorithm 1, Algorithm 2 presents a more detailed, low level version of the k-MS algorithm.

**Algorithm 2.** Low-level k-Morphological Sets algorithm.

**Data:** $B$ is the structuring element and $k$ is the desired maximum number of clusters to be created

1 **begin**
2     finished $\leftarrow$ false;
3     $\delta \leftarrow 1$;
4     **while** *!finished* **do**
5        idempotent $\leftarrow$ true; finished $\leftarrow$ true;
6        Reset or remove values from the kArray (the array contains at most $k$ different values);
7        **for** *every instance or data point $p$ in the input dataset* **do**
8           $p_{aux} \leftarrow p$;
9           If the surroundings of pixel $p$, respecting the structuring element $B$ and $\delta$, contains a higher value than $p$, then $p$ receives the highest surrounding value;
10          **if** *!idempotent* **then continue**;  // `first early break`
11          **if** $p_{aux} \neq p$ **then** idempotent $\leftarrow$ false;
12          **if** *!finished* **then continue**;  // `second early break`
13          **if** *kArray does not contain the value of $p$* **then**
14             **if** *kArray is fully populated* **then** finished $\leftarrow$ false;
15             **else** Add the value of $p$ to kArray at a vacant position;
16        **end**
17        **if** *idempotent* **then** $\delta \leftarrow \delta + 1$;
18        **else** $\delta \leftarrow 1$;
19        finished $\leftarrow$ finished **and** idempotent;
20     **end**
21     Returns all the instances $p$ where each one is indexed with a unique cluster index (amount of clusters $\leq$ k);
22 **end**

### 3.1. Computational complexity

As it can be seen in Algorithm 2, we have an inner "for" loop that depends on the total number of instances $n$ in the dataset. Within this loop, there is a loop that is dependent on the $k$ variable. Thus, in the worst case scenario, there is a total of $O(n \times k)$ steps for computing a single dilation of the dataset. The maximum number of dilations for two clusters to reach each other in the worst case is equal to the distances of the two farthest instances in the dataset, divided by the initial $\delta$ (which is equal to 1 in our case).

Let us assume that $\hat{d}$ represents this maximum distance between the two farthest instances in the dataset. Then $\hat{d}/\delta$ represents the maximum number of dilations that can be performed for two clusters to reach one another. Since $k$ is the number of desired clusters and assuming the worst case scenario, there is only a single cluster to be produced by the algorithm regardless of $k > 1$ (this would not happen in practice, however it serves for the purposes of complexity analysis), then $\hat{d}/\delta$ dilations would have to be performed a total of $k$ times at worst case. This leaves us with the complexity shown in Eq. (11), which is equivalent to $(k\hat{d}/\delta) \times (nk)$.

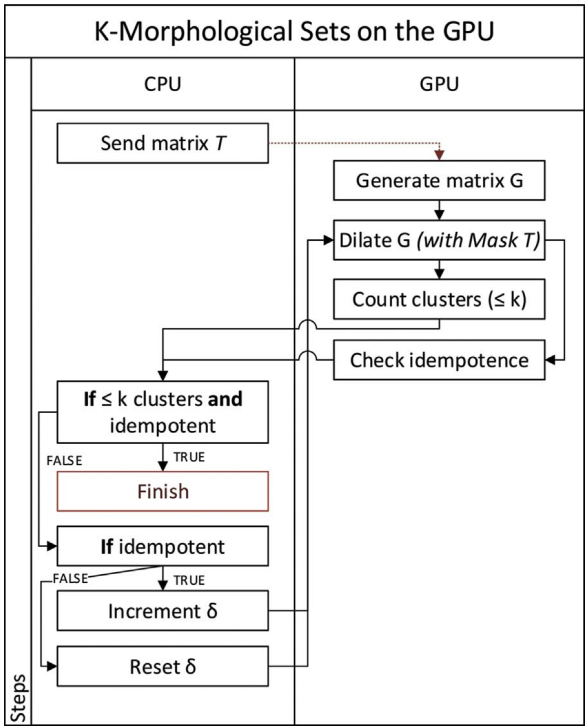

**Fig. 9.** Overall steps of the k-MS GPU implementation.

$$O\left(\frac{k^2 \hat{d} n}{\delta}\right) \quad (11)$$

Variables $\delta$ and $k$ are constants and independent of $n$ so they can be excluded from the analysis. Thus, this results in a $O(\hat{d}n)$ complexity. The variable $\hat{d}$ can be expressed in terms of $n$, since if the number of instances in the dataset increases, so can the maximum distance between two possible instances. In this highly improbable worse case scenario, we would have $O(n^2)$.

As a remark, Karypis et al. [18] do not provide a complexity or time analysis regarding Chameleon in their work. D. Liu et al. [19] and Yousri et al. [21] provide a complexity of $O(n \log n)$ for the TRICLUST and Mitosis algorithms. Although the worst case complexities of their algorithms are better than ours, we can say that in practice our algorithm is faster, as discussed in Section 4.3.

### 3.2. Parallel and GPU aspects

Besides the CPU implementation of the algorithm, we propose and evaluate parallel implementations of k-MS, including a GPU implementation as well. The GPU algorithm was divided in two kernels. The first kernel takes the boolean matrix $T$ and generates $G$, and the other kernel performs the remaining operations. Fig. 9 illustrates the steps of the k-MS algorithm and its synchronization with the CPU.

The parallel versions of the algorithm (CPU and GPU) work very similarly to the one shown in Algorithm 2. The main difference is that the "for" loop in line 7 is processed in parallel. In the GPU implementation, the early breaks in lines 10 and 12 do not produce an improvement in the processing time, since GPUs use the SIMD (Single Instruction Multiple Data) paradigm, where the same instruction is performed in every thread at the same time. Thus, even if some threads escape from the loop early, other threads that are scheduled to run together will eventually push the processing time to the worst case, as if the early break instructions are not present.

Contrariwise, the early breaks do work in the CPU parallel implementations because they do not follow the SIMD paradigm. In fact, the cores can process different instructions in parallel. It is also possible to infer that the performance of the GPU algorithm will not be good for large values of $k$. This happens because of two facts: (1) the early breaks do not work, and (2) more atomic operations have to be performed in each position of the k-array to avoid concurrency of the various threads, and performing a slow operation several times adds a significant overhead to the running time.

The cluster counting in the GPU is performed as follows: after the dilation of each pixel in $G$ (line 9 of Algorithm 2), a "for" iterates through an array of size $k$. If the value is not within the array and the array is not full, then the value is placed at a vacant position using the *atomic compare and swap* operation. Otherwise, if the value is not within the array and the array is full, then an *atomic and* sets the *finished* boolean variable to false. Similarly to the sequential version, the output of the algorithm is the $G$ matrix with $k$ or less distinct indexes or grey level values (plus the background value $S$) as well as the $k$-array, which contains every index of every cluster in the matrix. Each cluster has a unique and distinct index with relation to the remaining ones.

### 3.3. Memory consumption

The k-MS algorithm requires $g(n, L, k)$ bits of memory, as shown in Eq. (12), where $n$ represents the number of dimensions of the instances, $L$ represents the amount of instances and $k$ stands for the size of the $k$-array, which is also the number of desired clusters. We have implemented the algorithm using integers of 32 bits. However, this can be increased or decreased as desired, depending on the input dataset.

$$g(n, L, k) = sizeof(int) \times (nL + k) + sizeof(\delta) + sizeof(finished)$$
$$+ sizeof(idempotent) \quad (12)$$

In our specific case, we have considered $\delta$ to be an integer so that $sizeof(\delta)$ was equal to 32 bits and $sizeof(finished) + sizeof(idempotent)$ was equal to 2 bits, each one being boolean variables. The experiments in this work were not sufficient to overflow the memory of the used GPU (Nvidia GTX 960M). Furthermore, the algorithm requires subtle amounts of local memory, which would not overflow virtually any GPU shared nor local memory.

## 4. Experimental results

Three distinct groups of experiments are performed in the following sections. The first group, described in Section 4.1, evaluates and compares the results obtained with k-MS and other popular clusterization algorithms that are usually employed in high dimensional clusterization and do not necessarily aim to be sensitive to density and shapes. The results of the clusterizations are analysed by volunteers, who were instructed to select the most human-like segmentations. We also analyse how the k-MS algorithm visually behaves as the variable $k$ is increased. The dataset used in this section is noiseless.

The second group of experiments, addressed in Section 4.2, compares the visual results obtained in a noisy publicly available dataset [18]. In this case, we compare k-MS to clusterization algorithms with similar intents. The behavior of the k-MS algorithm with distinct input parameters is also analysed. Finally, in the third group of experiments, in Section 4.3, we perform an extensive time analysis, comparing the running times obtained by the different implementations of k-MS with the other clusterization algorithms. The structuring element used for almost this entire section was the one shown in Fig. 6.

### 4.1. Noiseless morphology recognition experiment

For this group of experiments, we consider the input data shown in

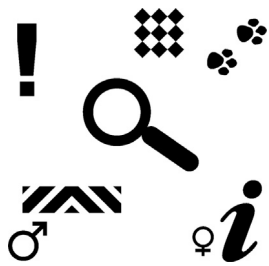

**Fig. 10.** The input *T* matrix.

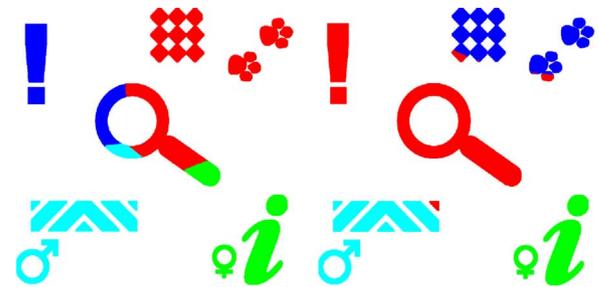

(a) SimpleKMeans  (b) EM

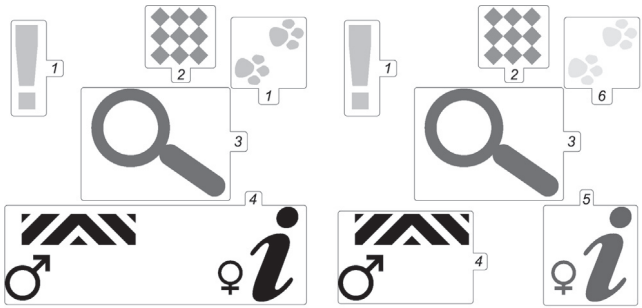

(a) $k = 4$  (b) $k = 6$

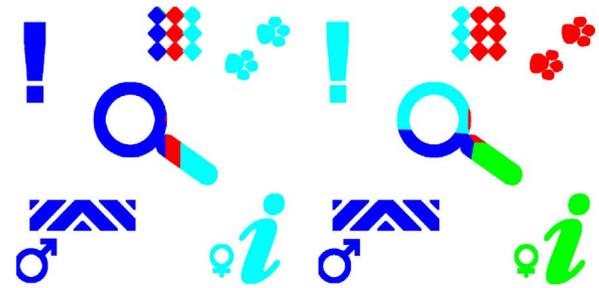

(c) MTree  (d) Farthest First

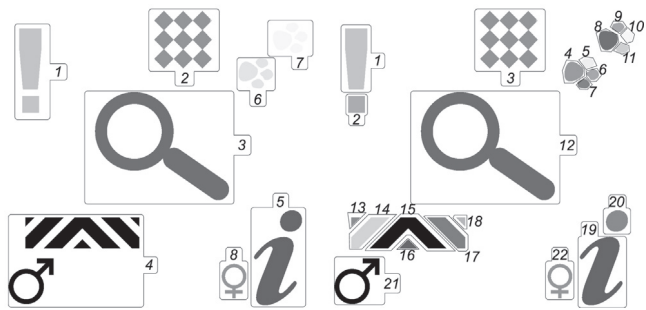

(c) $k = 8$  (d) $k = 25$

**Fig. 11.** Visual results of k-MS with varying values of $k$.

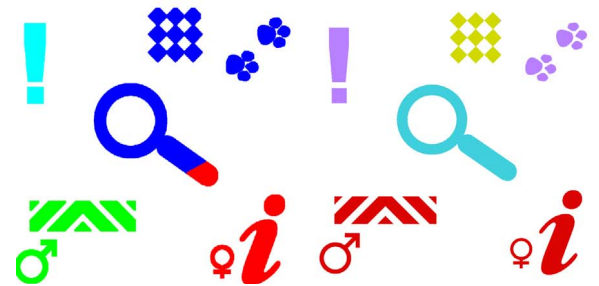

(e) Canopy  (f) k-MS

**Fig. 12.** Comparison of the visual results of k-MS to other clusterization algorithms for $k$=4.

Fig. 10, where the black pixels represent the instances in a 2 dimensional space. The first experiment consists of running the k-MS algorithm while varying the input $k$ variable, in order to show how the algorithm would group the different objects in the image gradually. Fig. 11 shows the results of the clusterizations for different values of $k$.

It is possible to see in Fig. 11-(a) that, for $k$=4, the k-MS algorithm grouped the exclamation mark and the paws in the same cluster. This occurred because the bordering limits of the image were disregarded, so that if the position $p$ of the pixel is larger than the *width* of the image then the reminder of the division of $p$ by *width* ($p\%width$) is considered instead. For $k$=6, shown in (b), the algorithm separated this cluster and the big cluster at the bottom that were previously grouped together for $k$=4. The amount of valid clusters keeps increasing until $k$=22. If $k > 22$, the algorithm still separates the dataset in 22 clusters, such as shown in (d), where $k$ was equal to 25.

In addition, we compare the clusterization results obtained by k-MS with two values of $k$ for various clustering algorithms that support the $k$ variable in the Weka machine learning framework [27], namely, the k-Means implementation called SimpleKMeans [28], Expectation Maximization (EM) [29], MTree [30], Farthest First [31] and Canopy [32]. Fig. 12 shows the results of clusterization for $k$=4 and Fig. 13 for $k$=8. The different clusters are represented in different colors in this case. The images are a little bit different due to the fact that one of the images was outputted by our algorithm and the remaining were taken directly from the Weka visualization tool.

We presented the clusterization results of Fig. 10 to 73 volunteers and instructed them to answer the simple question: "If you were to separate the image in $k$ groups, which image among the ones in Figs. 12 and 13 would be the closest to your separation?". Two questions were asked for $k$=4 and for $k$=8, separately. Table 1 compares the average choice of these participants in percentage. The error column shows the average rate of incorrectly classified instances in relation to the clusterization performed by our algorithm.

Given the results in Table 1, we can conclude that k-MS is the most "human-like" algorithm among the evaluated ones. For $k$=4, the percentage of volunteers that voted for k-MS was 58.3%, which is significantly lower than for the $k$=8 case (73.5%). We hypothesize that this was because in the case of $k$=4, the exclamation mark was placed in the same cluster as the paws at the opposite side and this may have prevented some volunteers from voting for k-MS on this occasion. In this group of experiments, we have enabled the dilations to extrapolate the image boundaries on purpose to check how the volunteers would vote. For instance, if we go to the right direction of the paws, we extrapolate the image boundary and get to the left border of the image, encountering the exclamation mark faster than any other shape in the image. And this is the exact reason why the exclamation mark and the paws are placed in the same cluster for $k$=4. In conclusion, the choice for k-MS was significantly higher than for the remaining algorithms in both occasions.

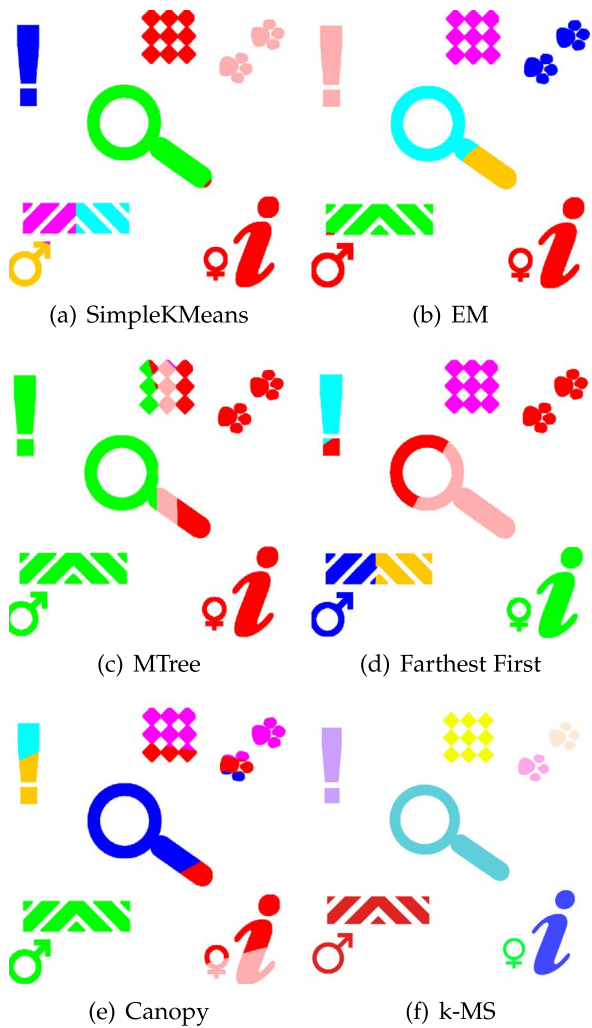

(a) SimpleKMeans  (b) EM

(c) MTree  (d) Farthest First

(e) Canopy  (f) k-MS

**Fig. 13.** Comparison of the visual results of k-MS to other clusterization algorithms for $k=8$.

**Table 1**
Comparison of obtained results based on human volunteers.

| Algorithms | Choice for k=4 (%) | Error for k=4 (%) | Choice for k=8 (%) | Error for k=8 (%) |
| --- | --- | --- | --- | --- |
| SimpleKMeans | 10.4 | 51.07 | 4.1 | 19.53 |
| EM | 6.3 | 37.46 | 16.3 | 20.22 |
| MTree | 4.2 | 59.05 | 4.1 | 59.02 |
| Farthest first | 12.5 | 50.49 | 2 | 27.07 |
| Canopy | 8.3 | 41.63 | 0 | 25.64 |
| **k-MS** | **58.3** | **0** | **73.5** | **0** |

### 4.2. Comparison with similar works

In this sub-section, we compare and analyse the clusterizations performed with k-MS, Chameleon, Mitosis, TRICLUST and M. Liu et al. algorithms on the dataset provided by Karypis et al. [18]. In contrast to the previous Section 4.1, the clusterization algorithms can deal with density and shapes and eventually recognize them. Fig. 14 shows the clusterizations obtained by each algorithm.

It can be seen in Fig. 14 that the clusterization obtained with the M. Liu et al. algorithm is a little difficult to visualize as the authors present it in their work. The symbols represent the different clusters and the letter Z in their image can be ignored. The authors of TRICLUST present the result without the noise of the initial dataset, so that we cannot determine how well their algorithm would handle cluster noise.

The clusterization of Mitosis and M. Liu et al. algorithms recognize exactly the right amount of genuine clusters, which are 9, and all noise in the image is included within these genuine clusters. This can be bad in some occasions such as when noise is expected to be removed or when it should not be aggregated in a genuine cluster.

On the other hand, the clusterization generated by Chameleon recognizes 9 genuine clusters as well as 3 other clusters that contain outliers. Chameleon and k-MS are the only algorithms capable of segregating outliers in different clusters. k-MS goes even further, it indicates whether the clusters can be formed given a predefined $k$ amount of clusters.

The result in Fig. 14-(e) shows the clusters found with k-MS algorithm, where each cluster is depicted in a different color. In noisy datasets such as this one, the value of $k$ should be large, otherwise k-MS would not be able to recognize the genuine clusters in the dataset. Furthermore, if the dataset is too sparse, it is also recommended to perform dilations on the dataset before applying k-MS, since the clusterization could be more accurate and converge faster.

If we consider low values for $k$, e.g., a maximum of 9 clusters, then the obtained result is as in Fig. 15-(a). This occurs because the instances are close to each other so that when they are dilated, several intersections occur, which would produce at the end of the processing the recognition of 9 clusters that are not the genuine ones. All genuine clusters were placed within a unique cluster in this case and the remaining were outliers (which can not even be seen properly because they are single points in the image).

If we increase the value of $k$ to 300, then we obtain the result shown in Fig. 15-(b). In this case, k-MS found 166 valid clusters, and the genuine clusters are now divided in only two different clusters. If we increase the $k$ to 450, then we obtain the result shown in Fig. 14-(e), where 450 valid clusters were found and the 9 genuine ones are separated correctly. In this case, k-MS algorithm took 6.33 s to generate these 450 valid clusters in the CPU.

The image shown in Fig. 14-(f) is a version of (e) where noise is removed. This removal is fairly simple to be performed. Given a threshold $\tau$, all the clusters that contain $\tau$ or less elements or pixels are erased. This is another particular and interesting property of k-MS. In the specific case of Fig. 14-(e), $\tau$ was equal to 200 pixels.

It is also possible to change the structuring element to obtain different results. For instance, instead of using $B_1 = \{(0, 0), (0, 1), (0, -1), (1, 0), (-1, 0), (-1, -1), (1, -1), (-1, 1), (1, 1)\}$ which is shown in Fig. 6, we could use for instance $B_2 = \{(0, 0), (0, 1), (0, -1), (1, 0), (-1, 0), (-10, -10), (10, -10), (-10, 10), (10, 10)\}$, where the diagonal pixels are more distant from the origin than others. For $k=450$ and $B_2$ we would have the result shown in Fig. 16. In this case, 372 valid clusters were found and the running time was at least 400 times faster than when considering $B_1$. Several combinations of clusters can be generated by tweaking the two parameters of k-MS.

At last, all the addressed algorithms were able to recognize the genuine clusters in this dataset. Due to this fact, we did not ask volunteers to choose a best segmentation from the available results, since the volunteers would be overly influenced by the different types of images that each author adopted.

### 4.3. Time analysis

At last, we perform an extensive time analysis experiment in this section. We used the GPU-oriented k-Means algorithm presented in [33], which is based on [34] and [35] for comparing and measuring the time performance of the algorithms on datasets with a large amount of instances. However, this algorithm could not handle the size of the instances we evaluate in this work, which is an advantage of our proposed technique. Therefore, our comparisons in this section address just the implementations of [34], which is a parallel k-Means on the CPU using OpenMP. We tested the sequential k-Means provided by them as well but it was always slower than the parallel one, so it was disregarded.

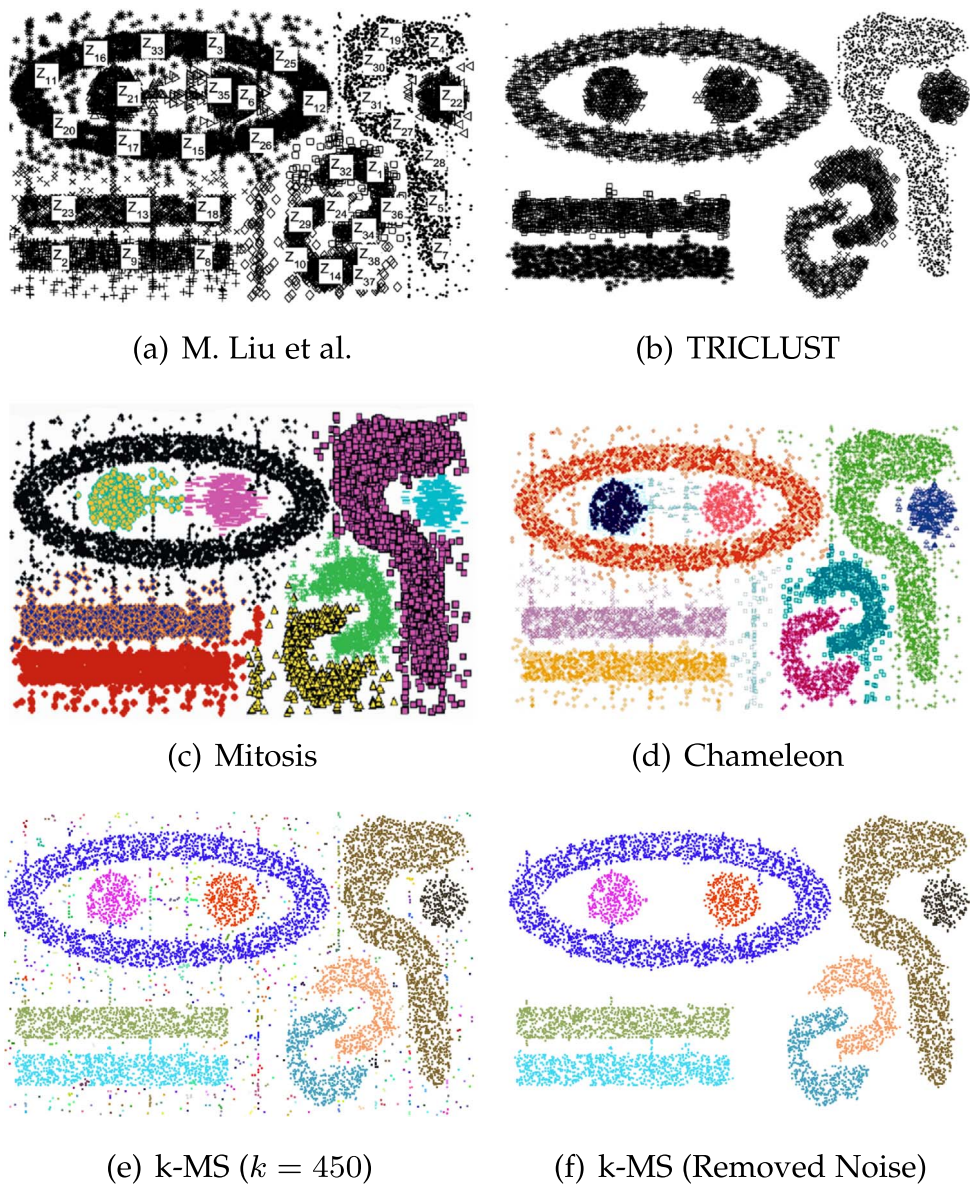

Fig. 14. Comparison of the visual results of k-MS and other clustering algorithms in finding genuine morphological clusters.

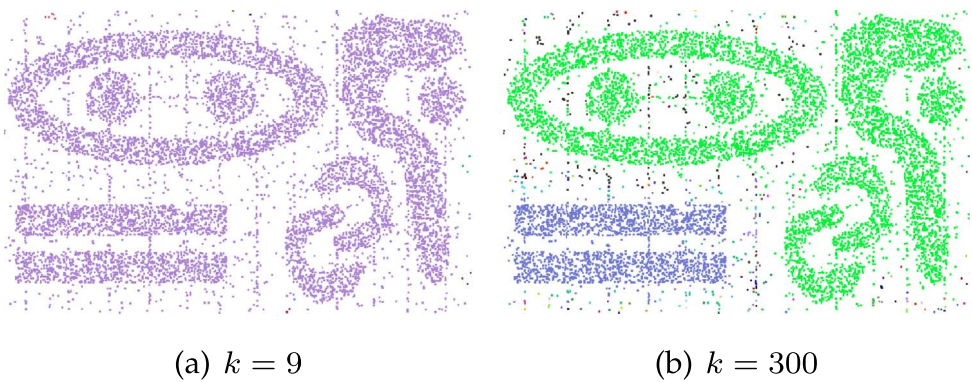

Fig. 15. Obtained clusters with lower values of $k$.

We set the threshold variable of the Liao algorithm [34] to $10^{-12}$, which essentially means that the algorithm stops when less than $10^{-10}$% of the instances change membership, i.e., change clusters. Since this is such a low number, we are essentially setting the algorithm to converge when the result is stable, i.e., when no instance changes membership. The comparisons are shown in Fig. 17, for instances of sizes: 512×512, 1024×1024 and 4096×4096, respectively. The presented results, however, would vary depending on the computer specification. In our experiments, we have used a Quad-core Intel i7-4720HQ, with 8 threads in OpenMP and in Java, and a Nvidia GTX 960M.

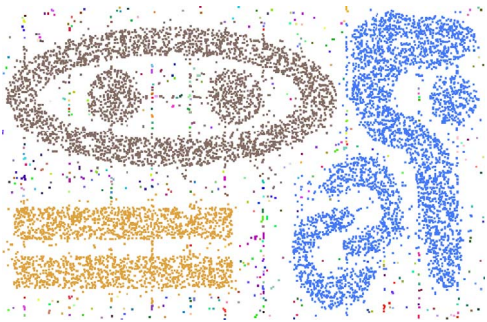

**Fig. 16.** Result for $k$=450 and $B_2$.

It is possible to see that k-Means starts faster for low values of $k$ but is overcome in every occasion after a particular size of $k$. As the size of $k$ increases, so does the running time of clusterization algorithms in general. However, due to the early break of the k-MS, the times obtained with k-MS may even decrease in the CPU as the value of $k$ increases. This happens because the clusterization converges faster with higher values of $k$ (less morphological reconstructions), and the early breaks reduce the time of the clusters verification. It is also possible to conclude that the GPU implementation is only viable if the amount of instances is very large and $k$ is small. Small values of $k$ reduce the amount of atomic operations that have to be performed, which reduces the processing time of the algorithm.

Table 2 shows the running times obtained with k-MS for smaller amounts of instances so that we can more appropriately compare k-MS to the remaining works. This is necessary because there are no reported analyses of the performances of these algorithms with a significantly high number of instances. The first column of the table indicates the values of $k$, and the first line indicates the amount of instances. In this case, just the CPU C/C++ sequential implementation was considered for a better and more fair comparison.

D. Liu et al. [19] provide a plot in their work, which contains the running times with different numbers of instances. Yousri et al. [21] provide four plots with the same intent, where each corresponds to a different dataset. These two works are the only ones (among [18–26]) that provide a time analysis while varying the amount of instances. We collected the approximate times provided in these plots, and compare them to the worst times that our algorithm achieved (the ones highlighted in red in Table 2). The comparison can be seen in Fig. 18.

However, the used datasets are different. In our case, we have worked with 512 × 512 images, where a total of $L$ instances were generated at random locations and $L$ is increased along the x axis. Yousri et al. [21] provide four plots for comparing the times, the ones with the lowest values were chosen, while in our case, the highest ones were selected so that we are already favouring Yousri et al. algorithm (Mitosis) as well as TRICLUST. Still, k-MS was faster than both Mitosis and TRICLUST in every occasion.

## 5. Conclusion

In this work, we propose and analyse a novel algorithm for clustering. The proposed algorithm is primarily based on morphological reconstructions, and we are the first to extend the reconstruction operation to a solid clusterization algorithm. According to the *no free lunch* theorem [36], there is no one model that works best for every optimization problem. The assumptions of a great model for one problem may not hold for another problem. Thus, it is common in machine learning to test multiple models and determine one that works best for a particular problem. However, beyond that, k-MS contains many particular and interesting properties.

We can fairly argue that our algorithm is substantially different

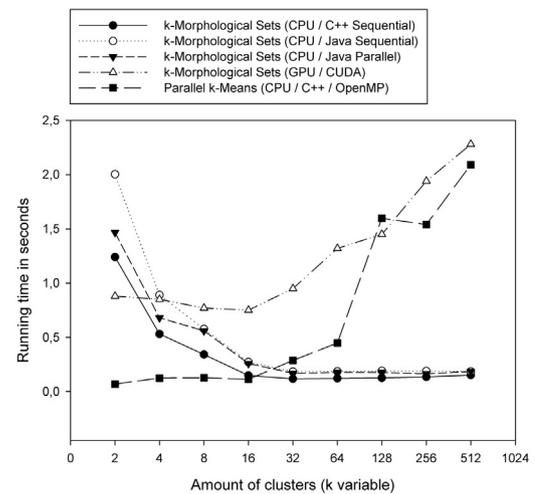

(a) 512 × 512 image, containing a total of 36,529 data points or instances.

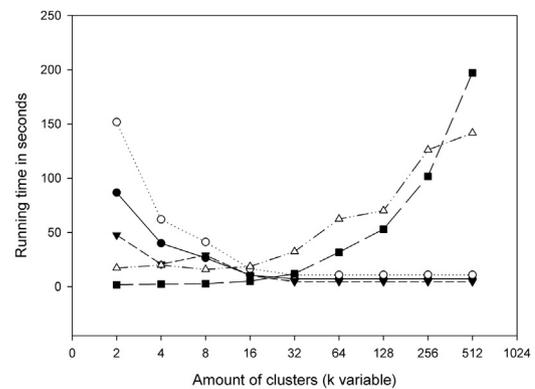

(b) 2048 × 2048 image, containing a total of 584,235 data points or instances.

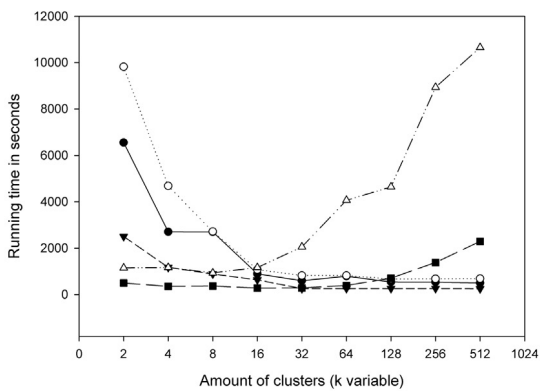

(c) 8192 × 8192 image, containing a total of 9,352,084 data points or instances.

**Fig. 17.** Time comparison between k-MS and a parallel k-Means considering a significant amount of instances.

from all the evaluated clusterization algorithms due to the fact that: (1) it has a sense of maximal clusters that can be created, (2) it is able to be efficiently implemented in parallel, (3) it permits the use of structuring elements, which can alter the way that clusters propagate and are created, (4) it provides a fast and simple noise removal-prone result and (5) it is also very simple. Apart from that, k-MS is also faster than all the remaining evaluated algorithms, including sequential and

Table 2
Run time performance (s) for the k-MS algorithm as a function of the number of instances and maximum cluster number $k$.

| k | 1000 | 2000 | 3000 | 4000 | 5000 | 6000 | 7000 | 8000 | 9000 | 10000 | 11000 | 12000 |
|---|---|---|---|---|---|---|---|---|---|---|---|---|
| 2 | 0.147 | 0.066 | 0.125 | **0.074** | **0.079** | **1.517** | **0.902** | **0.752** | **0.747** | **0.765** | **0.739** | **0.740** |
| 4 | 0.156 | 0.069 | 0.128 | 0.074 | 0.856 | 1.537 | 0.975 | 0.812 | 0.804 | 0.797 | 0.800 | 0.797 |
| 8 | 0.153 | 0.072 | 0.136 | 0.086 | 0.891 | 1.659 | 1.087 | 0.859 | 0.842 | 0.865 | 0.838 | 0.861 |
| 16 | 0.129 | 0.081 | 0.110 | 0.089 | 0.105 | 1.856 | 1.179 | 0.960 | 0.957 | 0.962 | 0.969 | 0.965 |
| 32 | 0.113 | 0.108 | 0.140 | 0.126 | 0.135 | 2.329 | 1.557 | 1.331 | 1.458 | 1.323 | 1.323 | 1.251 |
| 64 | 0.133 | 0.132 | 0.117 | 0.156 | 0.119 | 3.104 | 2.118 | 1.818 | 1.853 | 1.867 | 1.708 | 1.673 |
| 128 | 0.090 | 0.143 | **0.084** | 0.118 | 0.094 | 3.735 | 2.635 | 2.234 | 2.469 | 2.347 | 2.146 | 2.109 |
| 256 | **0.052** | 0.079 | 0.125 | 0.090 | 0.094 | 2.710 | 3.478 | 1.987 | 3.359 | 2.950 | 1.397 | 1.538 |
| 512 | 0.056 | **0.061** | 0.097 | 0.096 | 0.118 | 2.668 | 1.504 | 2.048 | 1.680 | 1.127 | 1.045 | 1.606 |

The times in bold represent the best times in the column, while the ones highlighted in red represent the worst.

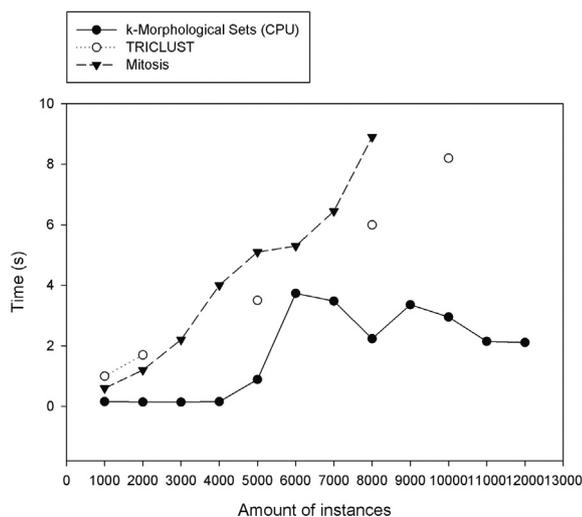

**Fig. 18.** Run times (s) comparison for a varying number of instances.

parallel k-Means, TRICLUST and Mitosis.

Although k-MS algorithm works for any $n$-dimensional dataset, we optimized the implementations used in this work for images. The implementations can be further extended to work with $n$-dimensional datasets and can be applied in several fields such as in data mining, machine learning and other areas of computer vision. Currently, the algorithm is more tuned towards image segmentation and visual pattern recognition tasks. As a final remark, all the source code and datasets used in this work are available in [37].

### Acknowledgements

The authors would like to thank CAPES, Brazil for the partial financial support in the form of scholarships.

### References

[1] A. Gionis, H. Mannila, P. Tsaparas, Clustering aggregation, ACM Trans. Knowl. Discov. Data 1 (2007).
[2] G. Gan, C. Ma, J. Wu, Data clustering: theory, algorithms, andapplications, Siam: Ser. Estat. Appl. Math. 20 (2007).
[3] J. Pena, J. Lozano, P. Larranaga, An empirical comparison of four initialization methods for the k-means algorithm, Pattern Recognit. Lett. 20 (1999) 1027–1040.
[4] A. Likas, N. Vlassis, J.J. Verbeek, The global k-means clustering algorithm, Pattern Recognit. 36 (2003) 451–461.
[5] R.S. Michalski, Knowledge acquisition through conceptual clustering: a theoretical framework and an algorithm for partitioning data into conjunctive concepts, Policy Anal. Inf. Syst. 4 (1980) 219–244.
[6] J. Goldsmith, Unsupervised learning of the morphology of a natural language, Comput. Linguist. 27 (2001) 153–198.
[7] L.F. Silva, A. Santos, R.S. Bravo, A.C. Silva, D.C. Muchaluat-Saade, A. Conci, Hybrid analysis for indicating patients with breast cancer using temperature time series, Comput. Methods Prog. Biomed. 130 (2016) 142–153.
[8] S. Goswami, L.K.P. Bhaiya, Brain tumour detection using unsupervised learning based neural network, Commun. Syst. Netw. Technol. (2013) 573–577.
[9] M. Su, C. Chou, A modified version of the k-means algorithm with a distance based on cluster symmetry, IEEE Trans. Pattern Anal. Mach. Intell. 23 (2001) 674–680.
[10] U. Fayyad, G. Piatetsky-Shapiro, P. Smyth, R. Uthurusamy, Advances in Knowledge Discoveryand Data Mining, AAAI Press, 1996.
[11] L. Torok, M. Pelegrino, D. Trevisan, E. Clua, A. Montenegro, A mobile game controller adapted to gameplay and user's behavior using machine learning, Entertain. Comput. – ICEC 2015 9353 (2015) 3–16.
[12] S. Koenig, R.G. Simmons, Unsupervised learning of probabilistic models for robot navigation, Proc. IEEE Int. Conf. Robot. Autom. 3 (1993) 1050–4729.
[13] J. Goutsias, H. Heijmans, Fundamenta morphologicae mathematicae, Fundam. Inform. 41 (2000) 1–31.
[14] J. Serra, Image Analysis and Mathematical Morphology, Academic Press, Inc., Orlando, FL, USA, 1983.
[15] L. Vincent, Morphological grayscale reconstruction in image analysis: applications and efficient algorithms, Image Process. 2 (1993) 176–201.
[16] F. Ortiz, F. Torres, Vectorial morphological reconstruction for brightness elimination in colour images, Real.-Time Imaging 10 (2004) 379–387.
[17] R.C. Gonzalez, R.E. Woods, S.L. Eddins, Morphological reconstruction: from digital image processing using matlab, URL ⟨http://mathworks.com/tagteam/64199_91822v00_eddins_final.pdf⟩.
[18] G. Karypis, E.H. Han, V. Kumar, Chameleon: hierarchical clustering using dynamic modeling, Computer 32 (1999) 68–75.
[19] D. Liu, G.V. Nosovskiy, O. Sourina, Effective clustering and boundary detection algorithm based on delaunay triangulation, Pattern Recognit. Lett. 29 (2008) 1261–1273.
[20] M. Liu, X. Jiang, A.C. Kot, A multi-prototype clustering algorithm, Pattern Recognit. 42 (2009) 689–698.
[21] N.A. Yousri, M.S. Kamel, M.A. Ismail, A distance-relatedness dynamic model for clustering high dimensional data of arbitrary shapes and densities, Pattern Recognit. 42 (2009) 1193–1209.
[22] A. Daneshgar, R. Javadi, S.B.S. Razavi, Clustering and outlier detection using isoperimetric number of trees, Pattern Recognit. 46 (2013) 3371–3382.
[23] K.M. Kumar, A.R.M. Reddy, A fast dbscan clustering algorithm by accelerating neighbor searching using groups method, Pattern Recognit. 58 (2016) 39–48.
[24] C. Zhong, X. Yue, Z. Zhang, J. Lei, A clustering ensemble: two-level-refined co-association matrix with path-based transformation, Pattern Recognit. 48 (2015) 2699–2709.
[25] M. Zarinbal, M.H.F. Zarandi, I. Turksen, Relative entropy collaborative fuzzy clustering method, Pattern Recognit. 48 (2015) 933–940.
[26] J. Yu, R. Hong, M. Wang, J. You, Image clustering based on sparse patch alignment framework, Pattern Recognit. 47 (2014) 3512–3519.
[27] M. Hall, E. Frank, G. Holmes, B. Pfahringer, P. Reutemann, I.H. Witten, The weka data mining software: an update, SIGKDD Explor. 11 (2009).
[28] D. Arthur, S. Vassilvitskii, k-means++: the advantages of carefull seeding, in: Proceedings of the eighteenth annual ACM-SIAM symposium on Discrete algorithms, pp. 1027–1035, 2007.
[29] T.K. Moon, The expectation-maximization algorithm, IEEE Signal Process. Mag. 13 (1996) 47–60.
[30] M.C. Mihǎescu, D.D. Burdescu, Using m tree data structure as unsupervised classification method, Informatica 36 (2012) 153–160.
[31] D.S. Hochbaum, D.B. Shmoys, A best possible heuristic for the k-center problem, Math. Oper. Res. 10 (1985) 180–184.
[32] A. McCallum, K. Nigam, L. Ungar, Efficient clustering of high dimensional data sets with application to reference matching, in: Proceedings of the sixth ACM SIGKDD internation conference on knowledge discovery and data mining ACM-SIAM symposium on Discrete algorithms, pp. 169–178, 2000.
[33] S. Giuroiu, Cuda k-means clustering, ⟨http://serban.org/software/kmeans/⟩.

[34] W. Liao, Parallel k-means data clustering, URL ⟨http://users.eecs.northwestern.edu/wkliao/Kmeans/index.html⟩.
[35] J. MacQueen, Some methods for classification and analysis of multivariate observations, Berkeley Symp. Math. Stat. Probab. 1 (1967) 281–297.
[36] D. Wolpert, W. Macready, Coevolutionary free lunches, IEEE Trans. Evolut. Comput. 9 (2005) 721–735.
[37] E.O. Rodrigues, k-morphological sets: Source and datasets, URL ⟨https://github.com/Oyatsumi/kMorphologicalSets⟩.

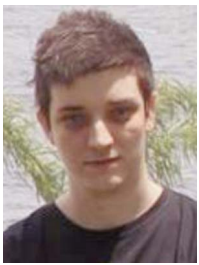

**Érick Oliveira Rodrigues** is currently a PhD student at Universidade Federal Fluminense (UFF), Niteroi, Brazil. He has received an award for the best master dissertation in exact sciences from UFF in 2016. In 2015, he received his master of science degree in visual computing from the same institution, and in 2013 he completed his undergraduate in computer systems, also from the same institution. He has been working mainly with computer vision, data mining and autonomous medical diagnosis in the last years. His main areas of interest are computer vision, artificial intelligence, GPU programming, optimization and bioinformatics.

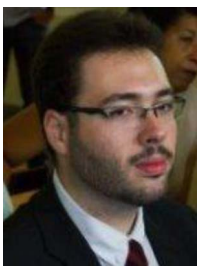

**Leonardo Torok** has a master degree in Visual Computing from Universidade Federal Fluminense (UFF), a degree in Computer Engineering from the Catholic University of Petropolis (2012) and is currently a PhD student in Visual Computing at Universidade Federal Fluminense, focusing on dynamic interfaces for game controllers using machine learning and mobile devices, covering the areas of gaming research, user experience, human-computer interface, mobile applications and data mining.

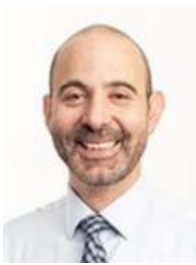

**Panos Liatsis** is a Professor and the Department Chair of Electrical Engineering at the Petroleum Institute in Abu Dhabi, UAE. Previously, he was a Professor of Image Processing and Head of Department of Electrical and Electronic Engineering at City, University of London, UK. His main research interests are in the areas of image and signal processing, machine learning and intelligent systems. Over the course of his career, he worked on developing tools and systems for analysis, control and decision support in diverse application areas, from autonomous vehicles to condition monitoring and biomedical imaging.

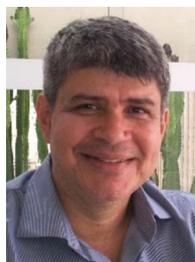

**José Viterbo** has a degree in Electrical Engineering (with emphasis in Computer Science) from the Polytechnic School of So Paulo University, a Masters degree in Computer Science, from Fluminense Federal University, and a PhD in Computer Science, from Pontifical Catholic University of Rio de Janeiro. Currently, he is Adjunct Professor at the Institute of Computing, Federal University Fluminense (IC/UFF). He coordinates the Laboratory of Real-Time Systems and Embedded Systems (LabTempo) and is an associate researcher at the Laboratory of Active Documentation and Intelligent Design (ADDLabs) and Laboratory of Management in Information and Communication Technology (GTecCom), at the same university. Also, he is a Director of Publications of the Brazilian Computer Society (SBC). He serves on the UFF Graduate Program in Computing (PPGC/UFF), where he conducts research in the area of ubiquitous computing, distributed inference, artificial intelligence, open data and efficient distributed data analysis.

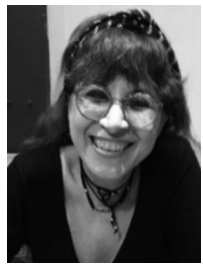

**Aura Conci** is full professor at Universidade Federal Fluminense (UFF) in computer science. In 1988, she received her PhD by Pontificia U. Catlica of Rio de Janeiro (PUC-Rio) in civil engineering. Post-doc in Universidad Rey Juan Carlos in 2011 in Madrid. Works with computer science ever since the creation of the computer science department in UFF. Her main research areas are computer vision, including mathematical morphology, image processing, biomedical modelling and diagnosis, serious games as well as applications of artificial intelligence.